\pdfoutput=1
\documentclass[11pt]{article}

\usepackage[preprint]{acl}

\usepackage{times}
\usepackage{latexsym}

\usepackage[T1]{fontenc}
\usepackage[utf8]{inputenc}

\usepackage{microtype}

\usepackage{inconsolata}

\usepackage{graphicx}
\usepackage{booktabs}
\usepackage{multicol}
\usepackage{multirow}
\usepackage{makecell}
\usepackage{comment}
\usepackage[most]{tcolorbox}
\usepackage{colortbl}
\usepackage{tcolorbox}
\usepackage{xcolor}

\usepackage{changepage}

\definecolor{cellgreen}{RGB}{208,239,215} % light green
\definecolor{cellred}{RGB}{246,215,215}   % light red

\newtcbox{\goodcell}{
  on line,
  colback=cellgreen,
  colframe=cellgreen,
  arc=3pt,
  boxsep=1pt,
  left=1pt,
  right=1pt,
  top=0.5pt,
  bottom=0.5pt,
  boxrule=0pt
}

\colorlet{cellgreenlight}{cellgreen!40!white}

\newtcbox{\secondcell}{
  on line,
  colback=cellgreenlight,
  colframe=cellgreenlight,
  arc=3pt,
  boxsep=1pt,
  left=1pt,
  right=1pt,
  top=0.5pt,
  bottom=0.5pt,
  boxrule=0pt
}

\newtcbox{\badcell}{
  on line,
  colback=cellred,
  colframe=cellred,
  arc=3pt,
  boxsep=1pt,
  left=1pt,
  right=1pt,
  top=0.5pt,
  bottom=0.5pt,
  boxrule=0pt
}

\title{Many Ways to Be Fake: Benchmarking Fake News Detection Under Strategy-Driven AI Generation}

\author{Xinyu Wang, Sai Koneru\thanks{These authors contributed equally.}, Wenbo Zhang\footnotemark[1], Wenliang Zheng\footnotemark[1],  Saksham Ranjan, Sarah Rajtmajer  \\
Pennsylvania State University\\
University Park, PA 16802, USA \\
\texttt{\{xzw5184,sdk96,wjz5120,wmz5132,sjr6223,smr48\}@psu.edu} \\
}

\begin{document}
\maketitle
\begin{abstract}

Recent advances in large language models (LLMs) have enabled the large-scale generation of highly fluent and deceptive news-like content. While prior work has often treated fake news detection as a binary classification problem, modern fake news increasingly arises through human–AI collaboration, where strategic inaccuracies are embedded within otherwise accurate and credible narratives. These mixed-truth cases represent a realistic and consequential threat, yet they remain underrepresented in existing benchmarks.

To address this gap, we introduce MANYFAKE, a synthetic benchmark containing 6,798 fake news articles generated through multiple strategy-driven prompting pipelines that capture many ways fake news can be constructed and refined. 
Using this benchmark, we evaluate a range of state-of-the-art fake news detectors. 
Our results show that even advanced reasoning-enabled models approach saturation on fully fabricated stories, but remain brittle when falsehoods are subtle, optimized, and interwoven with accurate information. 

\end{abstract}

\section{Introduction}
    
Commercial large language models (LLMs) like ChatGPT\footnote{https://chatgpt.com} and Copilot\footnote{https://copilot.microsoft.com} have democratized text generation, enabling not only productive applications but also the rapid production of deceptive content \cite{barman2024dark,shah2024navigating,pan2023risk}. Fake news, in particular, has evolved into a complex challenge that extends well beyond simple binary assessment. 
Modern fake news is often produced through deliberate, multi-stage human–AI collaboration, where prompting and iterative refinement are used to fabricate, distort, or selectively alter information to varying degrees \cite{wang2025have}.

Yet, most existing taxonomies and benchmark datasets treat fake news as a monolithic phenomenon. They are primarily topic-oriented and contain stories that are predominantly or entirely false \cite{shu2020fakenewsnet,wang2017liar, nanabala2024unmasking}. As a result, detection models evaluated on these benchmarks may report high accuracy yet fail to generalize to more realistic settings. In particular, prior work shows that both state-of-the-art LLM detectors and human evaluators achieve only about 60\% accuracy when assessing fake news created through human–AI collaboration \cite{wang2025have}. This mismatch arises because the most consequential AI-generated fake news is rarely fully fabricated, but instead embeds strategic inaccuracies within otherwise accurate, well-structured, and credible narratives—cases that are both more persuasive and systematically underrepresented in existing benchmarks.

To address this gap, we introduce \textbf{MANYFAKE}, a large-scale synthetic benchmark containing 6,798 fake news articles generated through strategy-driven human–AI prompting pipelines. In this dataset, \emph{fake news} is defined through the generation process itself--as news-like texts produced by LLMs through controlled generation strategies that seek to fabricate or distort narratives.

MANYFAKE explicitly captures different forms and degrees of falseness, ranging from direct fabrication to subtle fact distortion and stylistic imitation, as well as downstream optimization and refinement practices that improve fluency and plausibility. Using this benchmark, we evaluate a range of state-of-the-art fake news detectors, including LLM-based systems with and without explicit reasoning capabilities. Although MANYFAKE is fully synthetic, it is grounded in verified real-world claims and explicitly models human–AI generation strategies observed in practice, allowing controlled evaluation of realistic misinformation scenarios. 

Key contributions of this work are: (1) a strategy-driven taxonomy that characterizes many ways human–AI collaboration produces fake news with varying forms and degrees of falseness; (2) MANYFAKE, a synthetic benchmark constructed using this taxonomy to enable controlled evaluation across realistic generation strategies; and (3) empirical findings suggesting that even advanced reasoning-enabled LLM detectors approach saturation on fully fabricated stories, yet remain brittle to strategically optimized, mixed-truth fake news. 
\begin{figure*}[h!]
    \centering
    \includegraphics[scale=0.53]{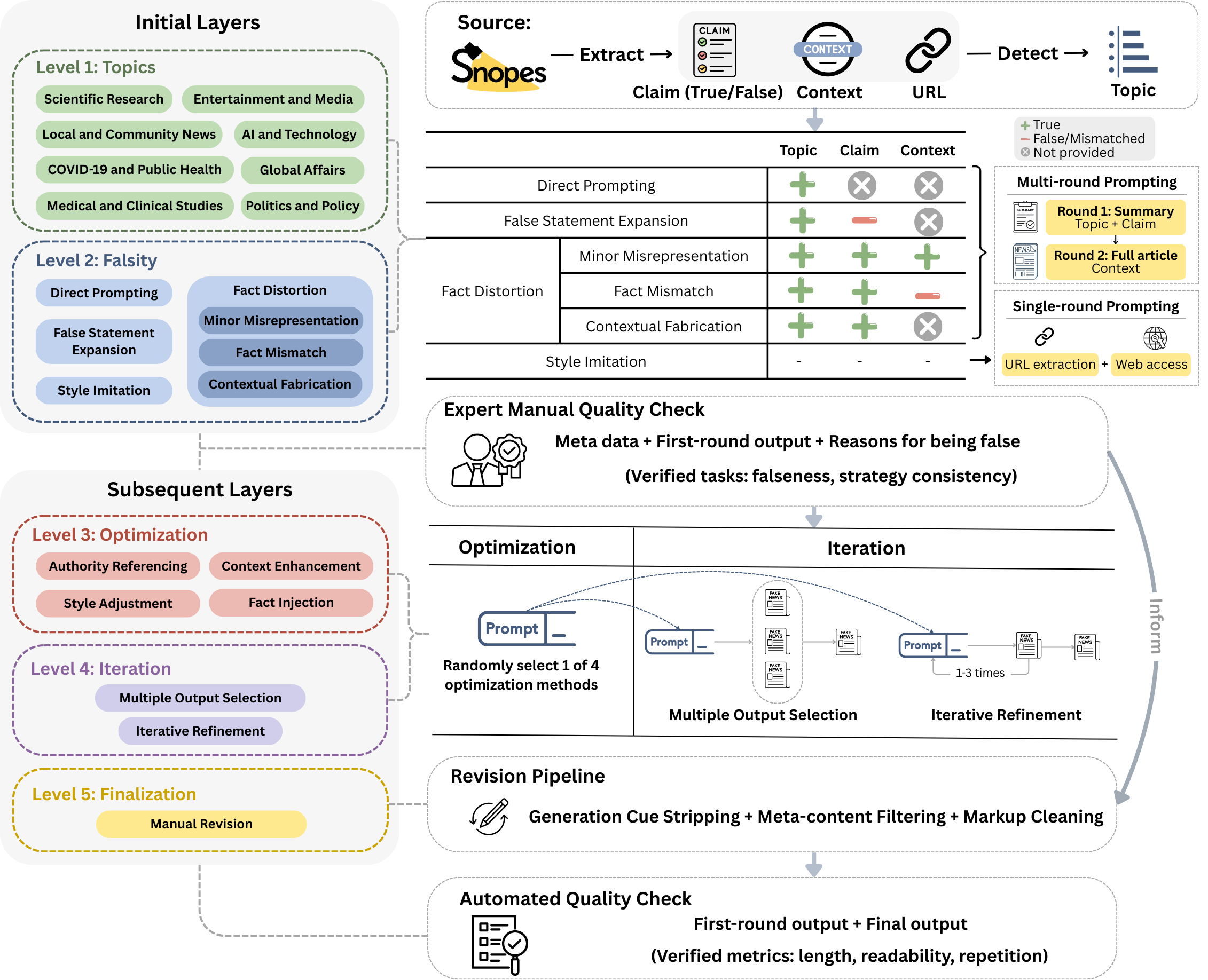}
    \caption{Framework for generating strategy-driven fake news. \textbf{Left}: 5-level taxonomy capturing topical domains and falsity strategies. \textbf{Right}: Operational pipeline, including claim and context extraction from Snopes, single- and multi-round prompting for generation, expert validation, optimization,  iteration, finalization, and final quality check. }
    \label{fig:taxonomy}
\end{figure*}

\section{Related Work}
\subsection{Taxonomies and benchmarks}% for fake news}

Prior work has characterized fake news through the lens of claim verifiability and intent to mislead readers \cite{allcott2017social, lazer2018science}, situating it within a larger ecosystem of information pollution ecosystems that distinguishes misinformation and disinformation \cite{wardle2017information}. While foundational, these frameworks were designed primarily for human-authored content. To empirically study these, prior work established foundational datasets like LIAR, which provides short political claims with fine grained truthfulness labels \cite{wang2017liar}, and FakeNewsNet, which curates news articles grounded in fact check sources and enriches them with social context signals to support broader modeling of misinformation in online ecosystems \cite{shu2020fakenewsnet}. However, these datasets do not fully capture the complexity and scalability of fake news generation enabled by LLMs. 

Newer corpora address this by incorporating LLM generated or LLM edited fake news. For example, FakeGPT uses LLMs to generate fake news and pairs this with detection and explanation\cite{huang2023fakegpt}, and explicitly operationalizes LLM-driven fake-news generation using a theory-motivated design \cite{wang2024megafake} or how AI-generated fake news competes with real news in recommendation systems \cite{hu2025llm}. Parallel to these resource developments, recent studies have focused on categorizing AI-generated fake news based on the types of prompts used, such as direct topic-based generation or the stylistic rewriting of existing articles \cite{chen2024llmgenerated}. While other works take intent, modality, generation methods into consideration \cite{yu2024fake}. Human-AI collaborative fake news generation studies suggest that the process includes a mix of multiple tactics to optimize deception, highlighting the need for taxonomies that reflect this multi-stage strategy driven creation \cite{wang2025have}.
%\subsection{Fake news benchmarks} %{\color{magenta}rename something like benchmarks?}
\subsection{Automatic fake news detection}
Fake news detection has progressed from early feature-based pipelines to transformer-era classifiers and, more recently, LLM-based detectors and reasoning-driven verification systems \cite{zhou2020survey,shu2017fake}. Early approaches relied on feature engineering using linguistic cues such as lexical \cite{rashkin2017truth}, sentiment \cite{potthast2017stylometric} markers from the news text to train traditional classifiers to distinguish between fake and real news. Other approaches involved supplementing with auxiliary information such as user comments \cite{shu2019defend, shu2017fake}. Subsequent deep learning work transitioned toward end-to-end modeling by learning representations from both content \cite{kaliyar2021fakebert} and context, including joint modeling of text with user-response patterns to capture temporal and social dynamics \cite{ruchansky2017csi}. Complementary directions improved robustness using multimodal and adversarial objectives to reduce event-specific artifacts \cite{wang2018eann}, developed unsupervised/weakly supervised formulations for label-scarce settings \cite{yang2019unsupervised}, and moved beyond shallow text cues by combining semantic representations with broader ecosystem signals \cite{shu2019beyond}. 

As LLMs are increasingly used to generate and rewrite news-like content \cite{ansari2026echoes}, detection models face new distributional shifts. Several studies show that AI-generated fake news can be substantially harder to detect than human-authored misinformation, especially when models are used to rewrite or style-transfer content to increase fluency and credibility \cite{chen2024llmgenerated,NEURIPS2019_3e9f0fc9}. Analysis on multi-domain datasets further shows that AI-generated fake news detectability varies substantially across domains and generation conditions \cite{nanabala2024unmasking}. In response, recent work evaluates LLMs directly as detectors via zero-shot or few-shot prompting with strong results on conventional datasets \cite{lucas2023fighting,pelrine2023towards,boissonneault2024fake}. Beyond end-to-end detectors, LLMs are increasingly used as modular components to provide intermediate signals, e.g., extract semantic features \cite{ma2024fake} or generate explanatory justifications \cite{wang2024llm} to be used in downstream detection. 

\section{Synthetic Dataset Generation}

Our proposed taxonomy and subsequent MANYFAKE dataset are grounded in the empirical findings reported in \citet{wang2025have}, systematically summarizing the human-AI collaborative strategies consistently employed during incentivized fake news creation.

\subsection{Taxonomy construction}

We model fake news generation as a structured, multi-stage process, decomposed into five levels: topics, falsity strategy, optimization, iteration, and finalization. This design reflects how human–AI collaboration incrementally constructs and refines deceptive narratives.

\noindent \textbf{Level 1: Topics.} To incorporate the content-driven approach of existing taxonomies, we consider 8 topics from \cite{wang2025have}: \emph{scientific research}; \emph{AI and technology}; \emph{local and community news}; \emph{COVID-19 and public health}; \emph{global affairs}; \emph{politics and policy}; \emph{medical and clinical studies}; and, \emph{entertainment and media}.\\ 

\noindent \textbf{Level 2: Falsity.} We build on four strategies observed in real-world AI-supported fake news generation \cite{wang2025have}.

\noindent \textit{\textbf{Direct prompting.}} Provide a clear, concise directive to the model, offering a constraint on the topic or style. Neither fact nor context are provided. 

\noindent \textit{\textbf{False statement expansion.}} Frame the story around a false statement, guiding the AI to produce content within that fabricated narrative. %{\color{magenta}We use a verifiably fake claim from the database and constrain with a topic input.} Neither context nor style are not provided.

\noindent \textit{\textbf{Style imitation.}} Provide an existing article; instruct the AI to create a similar but false story. Topic, factual claim, and corresponding context are all provided conceptually with embedded style. 

\noindent \textit{\textbf{Fact distortion.}} Provide true facts and prompt AI to generate fake news that incorporates and distorts these facts, blending truth with misinformation to create a convincing narrative. We further refine this strategy into three sub-strategies:

\begin{adjustwidth}{0.5cm}{0cm}
\noindent\textit{Minor misrepresentation.} Small, subtle alterations to factual details of the claim. 
\noindent\textit{Fact mismatch.} Combine fact with mismatched context to create a misleading narrative. 

\noindent\textit{Contextual fabrication.} Create new contexts that did not originally exist for the claim.\\
  
\end{adjustwidth}

\noindent \textbf{Level 3: Optimization.}
As observed in \citet{wang2025have}, a common practice in collaborative human-AI fake news generation is the refinement of AI-generated text using additional prompts. We simulate such secondary enhancements by prompting the model to revise its generated narrative using one of four strategies:

\noindent \textit{\textbf{Authority referencing.}}
Introduce fabricated citations, quotes, or attributions to authoritative or recognizable figures or institutions.

\noindent \textit{\textbf{Contextual enhancement.}}
Insert additional examples, details, or background elements that increase the depth or specificity of the narrative.

\noindent \textit{\textbf{Style adjustment.}}
Modify stylistic presentation, such as tone or structural organization, to better resemble journalistic conventions or  reporting styles.

\noindent \textit{\textbf{Fact injection.}}
Integrate supplementary true statements or factual elements to adjust the proportion of factual content within the story and increase its apparent plausibility.\\

\noindent \textbf{Level 4: Iteration.}
Iterative editing processes were commonly observed to refine fabricated content \cite{wang2025have}. Repeated revision or selection can improve fluency, coherence, and alignment with intended optimization cues. We distinguish two forms of iteration.

\noindent \textit{\textbf{Multiple output selection (MOS).}}
Multiple candidate variants are generated in parallel and a single version is selected as the final narrative.

\noindent \textit{\textbf{Iterative refinement (IR).}}
A single narrative is revised across multiple rounds; each iteration incorporates targeted adjustments.\\

\noindent \textbf{Level 5: Finalization.}
A final post-processing stage simulates human editorial intervention prior to dissemination. At this stage, the generated story is manually edited to remove artifacts associated with AI-generated text and to improve coherence. \\

The full set of prompts representing various strategy combinations are provided in Appendix~\ref{sec:appendix B}.

\subsection{Dataset collection and generation}
We constructed the dataset in two stages: \textbf{seed data collection} and \textbf{synthetic data generation}. First, we obtained verified claims and their associated contexts from the fact-checking platform \textit{Snopes.com}. Each claim was annotated with one of 8 topics to ensure thematic coverage. Building on this, we generated synthetic data by systematically combining different input components (topic, claim, and context). The following subsections describe these steps in detail.  

\subsubsection{Seed data collection and annotation}
\label{sec:seed-collection}

We sourced 4,000 (2,000 false and 2,000 true) seed facts from \textit{Snopes.com}, a fact-checking outlet established in 1994 and widely cited for debunking urban legends, rumors, and online misinformation. 

\paragraph{Source rationale.}
\textit{Snopes.com} publishes thousands of fact-checks pairing clearly-stated claims with editorial verdicts (\textit{true}, \textit{false}, \textit{mixture}) and accompanying justification grounded in primary sources, expert testimony, and original reporting. This structure provides (i) concise claim statements suitable for conditioning generation procedures and (ii) verified truth labels for benchmarking. 

\paragraph{Automated crawling and data processing.}
We implemented a robust, resumable crawler 
that targets the \emph{Snopes Fact Checks} archive\footnote{https://www.snopes.com/fact-check/} via the site’s ``latest'' pagination. 
For each fact-check, the crawler extracts a structured record (stored per-article) including the core meta data, claim and rating, article body, etc. 

For the \textit{core seed set}, we include entries with ratings that lie toward the ends of the rating spectrum (e.g., \textit{true} or \textit{false}, \textit{correct attribution}, \textit{mostly false}) and exclude vague categories such as \textit{mixture, unproven}, ensuring an unambiguous ground-truth signal for generation control and evaluation.

\paragraph{Topic assignment.}
We applied a two-stage topic labeling procedure. First, we assigned a topic to each instance based on \textit{title+context\_summary} using a zero-shot natural language inference model (\texttt{Qwen3-4B-Instruct-2507}), restricted to our eight topic categories. 
This step produced a single \textit{primary\_topic}, then provided as input to a second round of topic labeling on the output after finalization layer
using \texttt{GPT-4o-mini-2024-07-18}, which allowed us to capture potential topical drift. Topic assignment prompt template is provided in Appendix~\ref{sec:appendix B}.

\subsubsection{Synthetic corpus construction}
\noindent \textbf{Prompt design.} We developed two prompting pipelines to support synthetic news generation: direct prompting and multi-round prompting. In the direct prompting pipeline, all required fields are presented within a single instruction. Although simple, this configuration frequently activates model safety mechanisms and results in refusal responses. To reduce these failures, we implemented a multi-round prompting procedure informed by \cite{huang2023fakegpt}. In the first round, the model is provided with necessary fields and asked to outline how a news article would be written given topic and claim, yielding a concise 3–5 sentence summary. This summary is introduced as input to the second round, where the model is prompted to expand the outline into a full news article of $\sim$450–650 words. This approach substantially lowers refusal rates and reliably produces outputs that include all required fields. Detailed prompts are provided in Appendix~\ref{sec:appendix B}.\\

\noindent \textbf{Initial layers.}
Information provided to the model during generation follows the structure shown in Figure~\ref{fig:taxonomy}. While most strategies use the standard topic–claim–context inputs, two strategies require distinct procedures:

\noindent \textit{\textbf{Fact-mismatch construction.}}
In this strategy, the model receives a true claim paired with false context. 

To identify false yet topically-related context, we first generate concise summaries of all contexts in the dataset using \texttt{Qwen3-4B-Instruct-2507}.  
We then embed these summaries using \texttt{all-MiniLM-L6-v2} and apply approximate nearest neighbor retrieval to locate the most semantically-related but incorrect context from the corpus. This yields a context that maintains topical alignment with the claim while introducing a controlled factual inconsistency.

\noindent \textit{\textbf{Style imitation construction.}}
For style imitation, the model does not receive explicit topic, claim, or context fields. Instead, it is provided with the URL of an existing news article, which already contains the relevant narrative elements. The model is instructed to generate a new article following the writing style of the article from the URL. 
Since the original source encodes the topic, claim, and contexts, no additional fields are supplied for this category. We obtained URLs by scraping the reference links provided in Snopes articles, which indicate the original source of each news item. We retained only those sources that came from established news outlets rather than social media platforms to ensure that the writing style remained professional. \\

\noindent \textbf{Subsequent layers.}
Following the initial generation stage, we apply optimization, iteration, and finalization to produce refined synthetic news articles. These layers operationalize the conceptual processes defined in the taxonomy. 

\noindent \textit{\textbf{Level 3: Optimization.}}
For each generated datapoint, we randomly select one optimization operation from Authority Referencing, Contextual Enhancement, Style Adjustment, and Fact Injection. The chosen operation is used consistently during refinement, either through IR or MOS.

\noindent \textit{\textbf{Level 4: Iteration.}} For MOS, the model generates three optimized variants in parallel using the selected Level-3 operation and then selects the variant it judges to be the most convincing. For IR, the selected Level-3 operation is applied in sequence, with each application producing an updated version. The process repeats a random number of times between one and three, which introduces controlled variation and reflects the differing degrees of iterative editing that humans typically apply when refining text.

\noindent \textit{\textbf{Level 5: Finalization.}} We apply a fixed rule-based cleaning pipeline to the refined outputs, removing auxiliary sections (e.g., explanations), stripping metadata fields, and removing explicit synthetic or fictional disclaimers. All transformations are applied deterministically and do not modify the factual structure or narrative content.\\

\noindent \textbf{Model configuration.}
All data points were generated using the \texttt{gpt-5-mini-2025-08-07} model in batch mode with default parameters. Style imitation required web-access capabilities that this model does not support, so we used \texttt{Gemini-2.5-Flash} for that category. For both models, the max-completion-tokens parameter was set to 4,000 to support the production of full-length articles. For optimization layers, we used \texttt{Qwen3-4B-Instruct-2507}. 

\subsubsection{Corpus quality assurance}

\noindent \textbf{Expert verification.}
We performed expert verification immediately after the initial prompting stage. Two trained researchers independently reviewed each output as a quality control step, assessing whether it exhibited clear factual or misleading issues relative to the intended generation objective. Only suitable samples advanced to subsequent refinement stages, and the results informed post hoc cleaning by identifying outputs to discard and segments with strong model artifacts.

Each researcher reviewed 480 samples, corresponding to 80 news for each of the 6 generation strategies. Annotators followed a structured rubric that defined all  strategies, outlined general verification principles, and specified strategy-level criteria. They were permitted to consult reliable external fact-checking sources as needed. The two annotators achieved 93.5\% agreement. In total, 49 items were flagged during verification. Among them, 7 cases involved model refusals to generate non-factual content, 7 cases pointed out that it is fictional, 17 cases corrected the intended false statement, and 18 cases produced news that could not be confirmed as true or false. Based on these observations, we removed all items in the first three categories by applying strict keyword-based filtering followed by LLM-assisted filtering to the entire dataset. After data removal, the sampled data quality pass rate is 96\%.\\

\begin{figure}[h!]
    \centering
    \includegraphics[scale=0.42]{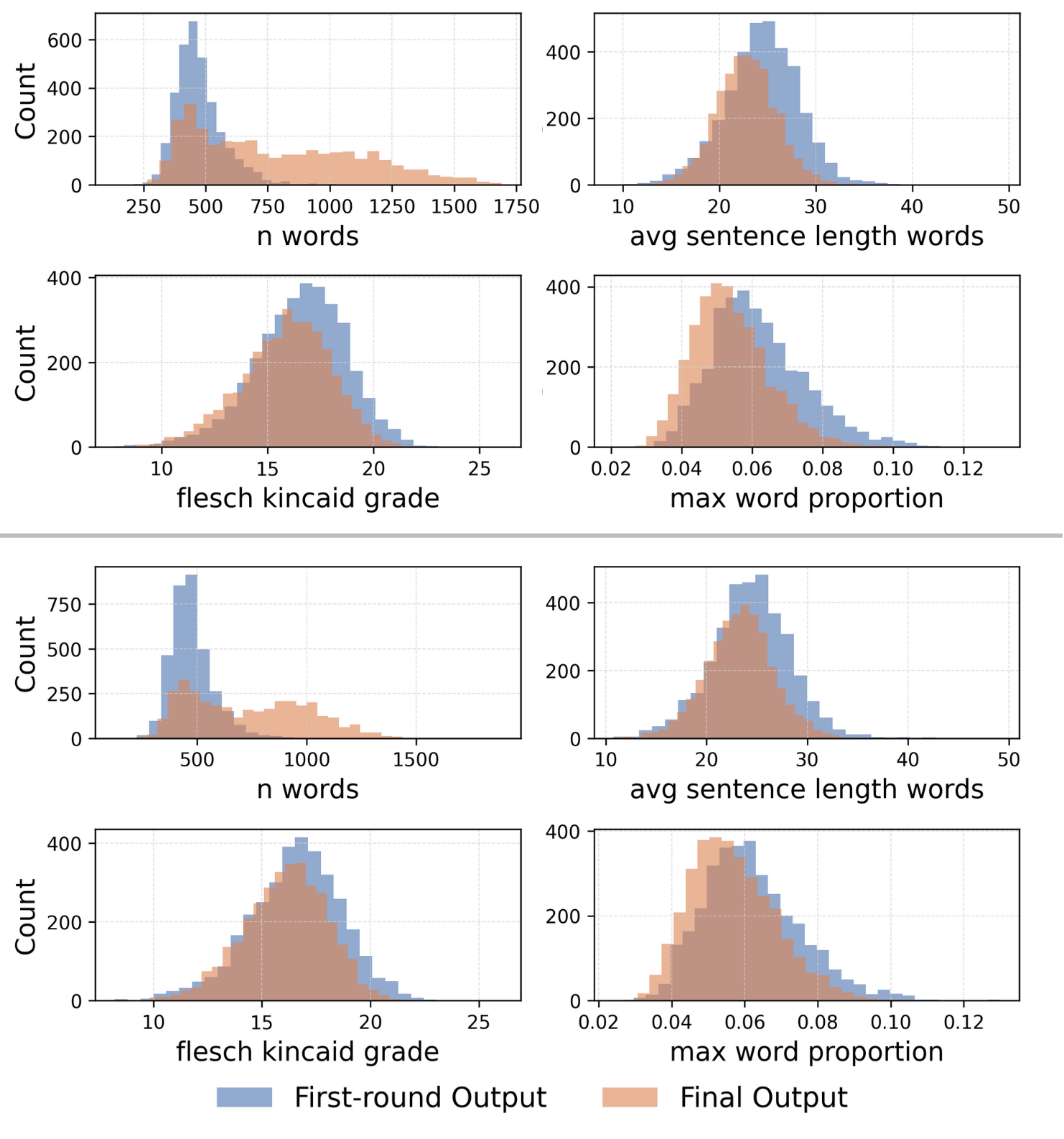}
    \caption{Distributions of automated verification metrics for first-round and final outputs. \textbf{Top}: IR;  \textbf{Bottom}: MOS. }
    \label{fig:distribution}
\end{figure}

\noindent \textbf{Automated verification.}
Alongside expert review, we developed an automated verification pipeline to screen both first-round and final outputs. We consider four metrics: \textit{word count}; \textit{average sentence length}; \textit{readability}; and, \textit{repetition}. 
Let $N_w$ denote the number of word tokens extracted using a word-level regex tokenizer, and let $N_s$ denote the number of sentences obtained by splitting on sentence-ending punctuation. Average sentence length is computed as $\overline{L}_{\text{sent}} = N_w / \max(N_s,1)$. Readability is measured using the Flesch--Kincaid grade \cite{kincaid1975derivation}, defined as $\mathrm{FKG} = 0.39(N_w/N_s) + 11.8(N_{\text{syll}}/N_w) - 15.59$, where $N_{\text{syll}}$ is the total number of syllables estimated via vowel-group counts. Repetition is quantified by the maximum word proportion $p_{\max} = \max_w \#(w)/N_w$, where $\#(w)$ is the frequency of word type $w$. Larger values of $\mathrm{FKG}$ and $p_{\max}$ indicate lower readability and higher repetition, respectively.
As shown in Figure~\ref{fig:distribution}, the distribution of $N_w$ peaks around $450$ words and exhibits a long right tail due to different optimization and iteration combinations, while the remaining metrics show approximately normal distributions without notable outliers, suggesting consistent generation quality without obvious irregular patterns.\\

\noindent \textbf{Final corpus composition.}
We constructed the final corpus by downsampling each topic to the same minimum count across strategies, removing topic-level imbalance from subsequent analyses. This approach ensures that any observed differences across strategies are not confounded by uneven topic representation.
The resulting dataset contains 6,798 items, with 3,390 from the IR iteration and 3,408 from the MOS iteration. Table~\ref{tab:example} in Appendix~\ref{sec:appendix A} shows an example datapoint, and Figure~\ref{fig:stats} in Appendix~\ref{sec:appendix A} summarizes the corpus composition. Topic distribution is identical across strategies; optimization operations are approximately uniformly represented, each accounting for $\sim$25\% of the data.

\section{Experiment Results}
\begin{table*}[t!]
\tiny
\centering

\begin{tabular}{llcccccc}
\toprule
\multirow{3}{*}{\textbf{Model}} & \multirow{3}{*}{\textbf{Iteration}} &
\multirow{3}{*}{\textbf{\makecell{Direct Prompts}}} &
\multirow{3}{*}{\textbf{\makecell{False Statement\\ Expansion}}} &
\multicolumn{3}{c}{\textbf{Fact Distortion}} &
\multirow{3}{*}{\textbf{\makecell{Style Imitation}}} \\
\cmidrule(lr){5-7}
& & & &
\textbf{\makecell{Minor Misrepresentation}} &
\textbf{\makecell{Fact Mismatch}} &
\textbf{\makecell{Contextual Fabrication}} &
\\
\midrule
\multirow{2}{*}{\makecell{
MistralLite-7B-GGUF}}
  & IR  & \badcell{1.24\%}& \badcell{4.96\%} & \badcell{0.53\%} & \badcell{0.53\%} & \badcell{0.71\%} & \badcell{3.36\%} \\
  & MOS & \badcell{2.29\%} & \badcell{4.05\%} & \badcell{0.70\%} & \badcell{0.35\%} & \badcell{0.88\%} & \badcell{5.81\%} \\
\midrule
\multirow{2}{*}{\makecell{Qwen2.5-7B-Instruct}}
  & IR  & 9.20\% &16.28\% & 1.77\% & 2.12\% & 2.83\% & 11.15\% \\
  & MOS & 9.51\% & 16.02\% & 2.46\% & 2.64\% & 2.11\% & 13.20\% \\
\midrule
%\multirow{2}{*}{Phi-3 (3.8B)}
%  & IR  &  &  &  &  &  &  \\
%  & MOS &  &  &  &  &  &  \\
%\midrule
\multirow{2}{*}{\makecell{
Gemma-3-4B-it}}
  & IR  & 14.69\% & 10.44\% & 6.55\% & 3.36\% & 5.84\% & 15.75\% \\
  & MOS & 16.02\% & 8.27\% & 6.51\% &2.82\% & 4.40\% & 21.65\% \\
\midrule
\multirow{2}{*}{\makecell{Llama-3.1-8B-Instruct}}
  & IR  &22.65\%  & 33.27\%&  6.02\%&  6.02\%&  5.31\%& 18.58\% \\
  & MOS &  19.72\%&  24.47\%&  4.93\%&  3.52\%&  3.70\%&21.30\%  \\

%\midrule
%\multirow{2}{*}{Deepseek-v2 (16B)}
%  & IR  &  &  &  &  &  &  \\
%  & MOS &  &  &  &  &  &  \\
\midrule
\midrule
\multirow{2}{*}{GPT-4o-mini-2024-07-18}
  & IR  & 12.74\%&18.94\%&3.36\%&1.42	\%&1.59\%&13.81\% \\
  & MOS & 13.38\%&17.61\%&3.52\%&3.17\%&1.58\%&12.68\% \\
\midrule

\multirow{2}{*}{GPT-4o-2024-11-20}
  & IR  &10.44\% &29.20\%&1.59\%&1.77\%&1.59\%&12.39\% \\
  & MOS & 11.62\%&28.87\%&1.41\%&2.29\%&1.23\%&13.20\% \\
\midrule
% \multirow{2}{*}{GPT-4o-mini-2024-07-18}
%   & IR  & 12.74\%&18.94\%&3.36\%&1.42	\%&1.59\%&13.81\% \\
%   & MOS & 13.38\%&17.61\%&3.52\%&3.17\%&1.58	\%&12.68\% \\
% \midrule
% \multirow{2}{*}{Gemini-2.0-Flash-001}
%   & IR  & 15.40\% & 24.07\% & 2.65	\% & 4.07\% & 3.54\% & 18.58\% \\
%   & MOS & 17.61\% & 23.24\% & 2.46\% & 3.87\% & 2.65\% & 17.43\% \\

% \midrule
\multirow{2}{*}{\makecell{GPT-5.1-2025-11-13\\None Reasoning Effort}}
  & IR  &  73.45\%& 86.73\% &48.50\%  &50.09\%  & 	59.65\% &61.59 \% \\
  & MOS &  70.60\%& 88.73\% & 40.85\% &45.95\%  &  50.53\%&  58.80\%\\
\midrule
\multirow{2}{*}{Gemini-2.0-Flash-001}
  & IR  & 15.40\% & 24.07\% & 2.65	\% & 4.07\% & 3.54\% & 18.58\% \\
  & MOS & 17.61\% & 23.24\% & 2.46\% & 3.87\% & 2.65\% & 17.43\% \\
\midrule
\multirow{2}{*}{\makecell{Gemini-3-Flash-Preview\\Minimal Thinking Level}}
  & IR  &  \secondcell{81.95\%}& \secondcell{91.68\%} &\secondcell{55.22\% } &\secondcell{57.52\%}  & \secondcell{66.55\%} & \secondcell{66.90\%} \\
  & MOS &  \secondcell{78.87\%}& \secondcell{92.61\%} & \secondcell{48.94\%} &\secondcell{55.81\%}  &  \secondcell{57.04\%}&  \secondcell{63.38\%}\\
\midrule
\midrule
% \multirow{2}{*}{Claude Opus4}
%   & IR  &  &  &  &  &  &  \\
%   & MOS &  &  &  &  &  &  \\
% \midrule

\multirow{2}{*}{\makecell{GPT-5.1-2025-11-13\\Medium Reasoning Effort}}
  & IR  &  67.43\%& 87.61\% &51.68\%&47.26\%  & 56.46\% & 63.72\% \\
  & MOS &  64.44\%& 88.91\% & 46.13\% &42.78\%  &  45.95\%&  60.39\%\\
\midrule
\multirow{2}{*}{\makecell{Gemini-3-Flash-Preview\\Medium Thinking Level}}
  & IR  & \goodcell{91.50\%} &\goodcell{94.69\%}  &\goodcell{66.02\%}  &\goodcell{65.84\%}  &\goodcell{78.05\%}  & \goodcell{74.87\%} \\
  & MOS &  \goodcell{90.14\%}&  \goodcell{95.42\%}&  \goodcell{60.74\%}&  \goodcell{63.20\%}&  \goodcell{70.25\%}& \goodcell{76.58\%} \\

%\midrule
%\multirow{2}{*}{LLaMA-3 70B (CoT)}
%  & IR  &  &  &  &  &  &  \\
%  & MOS &  &  &  &  &  &  \\
%\midrule
%\multirow{2}{*}{\makecell{Qwen3\\(4B-thinking)}}
%  & IR  &  &  &  &  &  &  \\
%  & MOS &  &  &  &  &  &  \\
%\midrule
%\multirow{2}{*}{\makecell{Phi-4-mini\\reasoning-4B}}
%  & IR  &  &  &  &  &  &  \\
%  & MOS &  &  &  &  &  &  \\
% \midrule
% \multirow{2}{*}{DeepSeek-R1}
%   & IR  &  &  &  &  &  &  \\
%   & MOS &  &  &  &  &  &  \\
%\midrule
%\multirow{2}{*}{openai/gpt-oss-20B}
%  & IR  &  &  &  &  &  &  \\
%  & MOS &  &  &  &  &  &  \\

\bottomrule
\end{tabular}
\normalsize
\caption{Binary detection results across generation strategies using standard models and models with integrated reasoning capabilities.(Color codes: \goodcell{Highest} \secondcell{Second-highest} \badcell{Lowest})}
\label{tab:experiment}
\end{table*}

We evaluate the MANYFAKE dataset using LLM–based approaches, including direct detection and reasoning-embedded detection.

\subsection{Detection Performance Across Generation Strategies}
To ensure consistent evaluation and avoid prompt-induced variation, we adopt the binary classification prompt from \cite{yan2025debunk} across all benchmark models, as shown in Appendix~\ref{sec:appendix B} (Benchmark Prompt). 
Table~\ref{tab:experiment} compares standard models and models with integrated reasoning support across different strategic generation methods.
We observe that standard commercial models (e.g., GPT-4o, Gemini-2, etc.) and leading open-sourced models (e.g., Qwen-2.5, LLaMA-3.1, etc.) show limited effectiveness across all strategies. We also observe that more advanced models, such as GPT-5.1 and Gemini-3-Flash, provide noticeable performance gains across strategies. Overall, accuracy is lower for strategies focused on partial falseness mixed with facts, such as minor misrepresentation, fact mismatch, and contextual fabrication.

We then examine whether state-of-the-art LLMs can improve fake news detection when guided to use explicit reasoning. 
Models with integrated reasoning support, such as GPT-5.1 and Gemini-3-Flash in medium-reasoning mode, achieve higher accuracy than when operating with minimal reasoning. Their improvements are mainly observed in cases where the fabricated content is obvious and the core claim is entirely false. This suggests that model advances are pushing fake news detection close to relative saturation within the benchmark for cases that are easy to flag, but meaningful progress now depends on improving performance on stories that blend misinformation with accurate details, where difficulty remains high.

To better understand how different elements of fake news relate to detection difficulty, we analyzed the reasoning summaries produced by the best performing model (Gemini-3-Flash) to identify which signals contribute most to its decisions. We use \texttt{Gemini-2.5-Flash} to extract the Top 3 determinants driving each judgment, mapped to 7 categories: timeline, entities, sources, facts, style, context, and structure (see Reasoning Analysis Prompt in Appendix~\ref{sec:appendix B}). Figure~\ref{fig:radar} visualizes the distribution of these determinants for both correctly and incorrectly classified cases. We observe that factual inconsistencies and incorrect entities dominate the signals when the model makes correct predictions, while cues related to context and factual claims appear more frequently in cases where the models fail to correctly identify. These patterns suggest that models could benefit from improved reasoning about contextual details and subtle distortions beyond the central claim, highlighting promising areas for targeted improvement in future detection system design.

\begin{figure}[h!]
    \centering
    \includegraphics[scale=0.43]{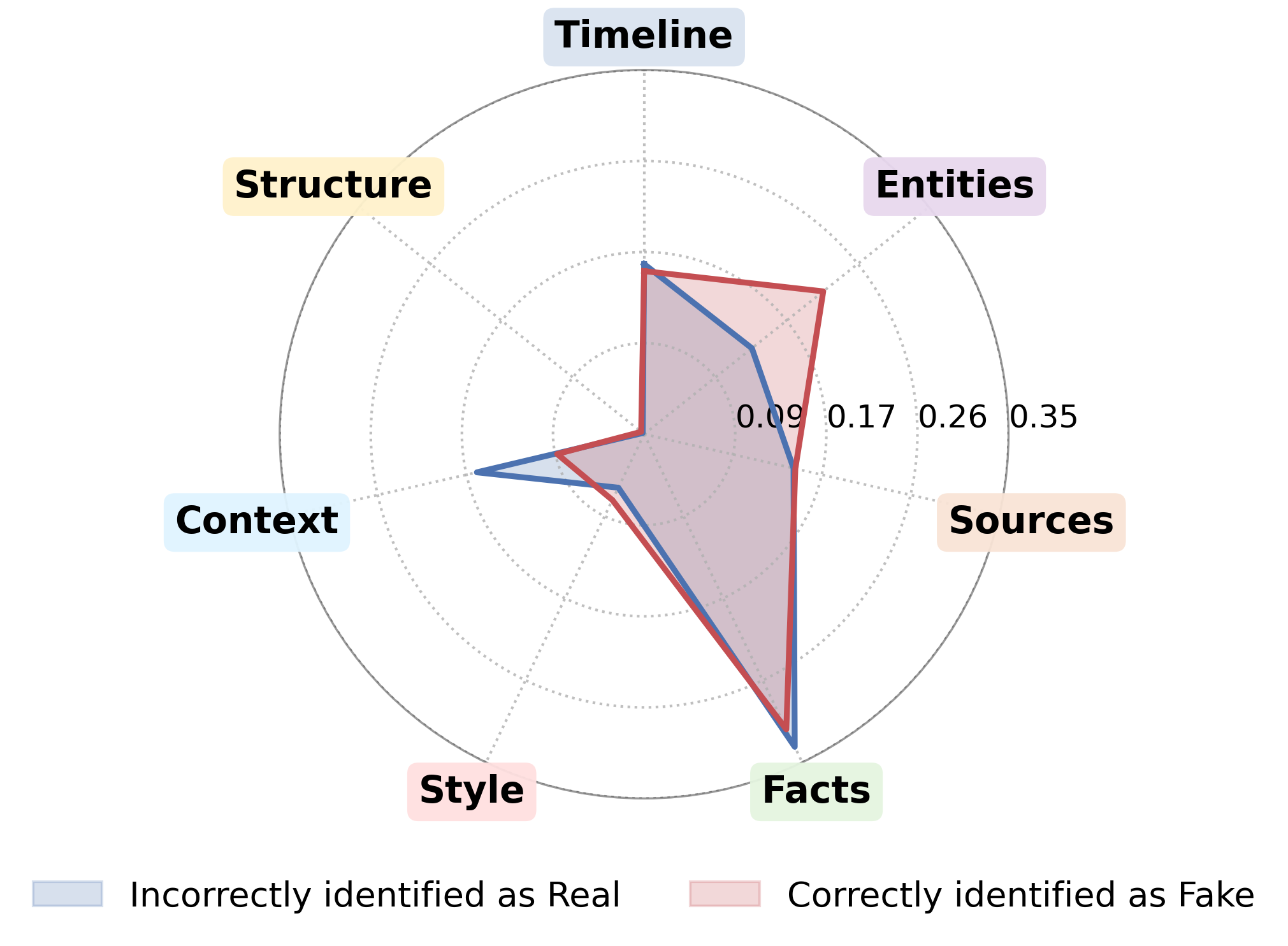}
    \caption{Radar plot showing the normalized distribution of dominant reasoning cues used in correct versus incorrect classifications (IR+MOS) by the best performing model (Gemini-3-Flash). }%{\color{magenta}font in legend (and numbers in plot) seems way too small}\textcolor{brown}{what about now?}
    \label{fig:radar}
\end{figure}

Taken together, these results point to two important observations. First, performance varies substantially across strategies, suggesting that high accuracy reported in earlier studies may come from datasets dominated by stories that are fully fabricated and easier to detect. Second, further progress in misinformation detection will depend on improving performance on mixed-truth cases that remain consistently difficult across models. These partially distorted stories are harder to identify and more reflective of real-world misinformation.

\subsection{Effects of Optimization and Topic on Detection Performance}
\begin{figure}[h!]
    \centering
    \includegraphics[scale=0.48]{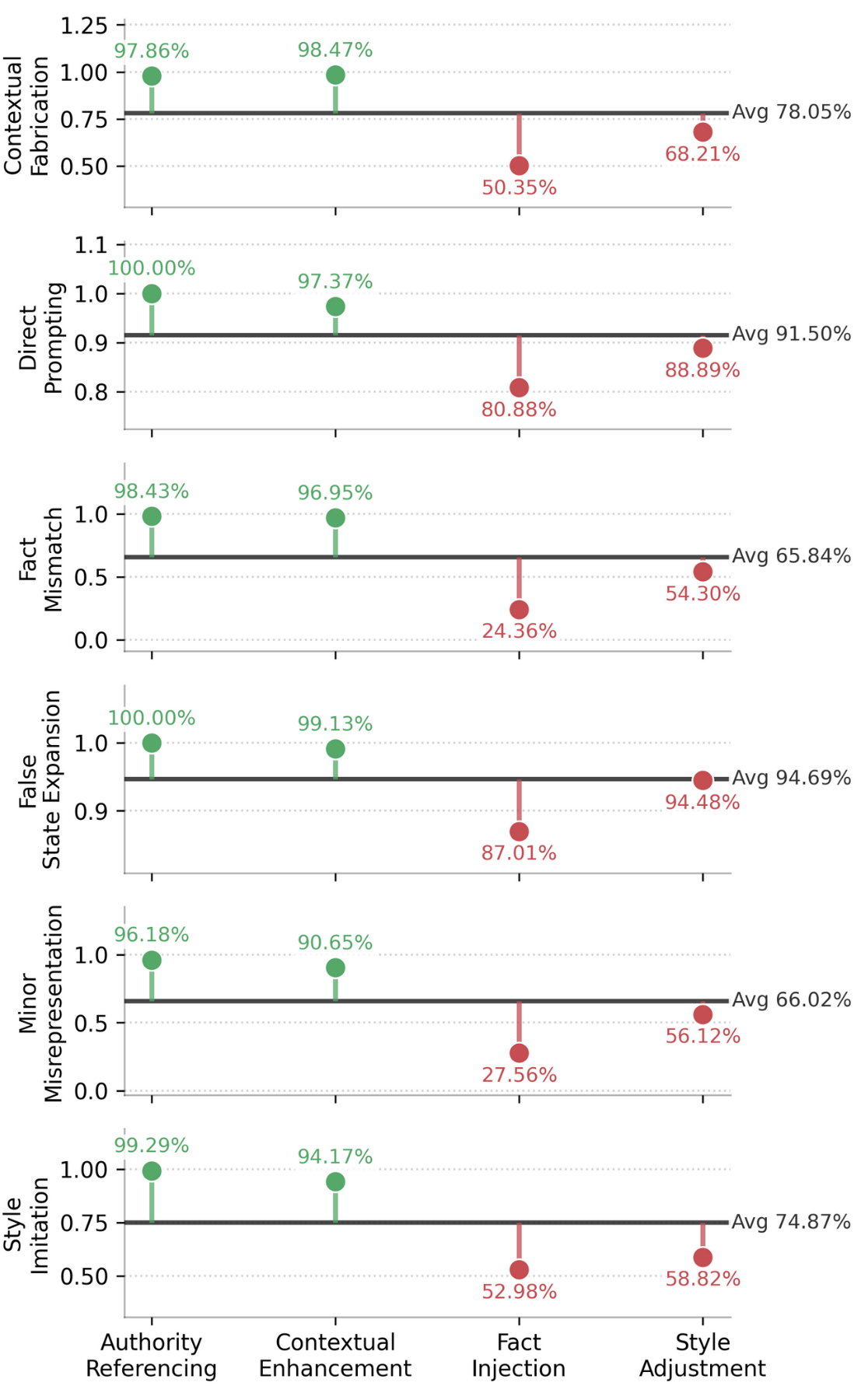}
    \caption{Detection accuracy across the interaction between generation strategies and Level-3 optimization operations for IR evaluated on Gemini-3-Flash.}  \label{fig:interaction}
\end{figure}
To understand how generation strategies and optimization operations jointly influence detectability, we evaluated all combinations of the six primary strategies with the four Level-3 optimization operations.

Figure~\ref{fig:interaction} reports accuracy stratified by the interaction between Level-3 optimization operations and each generation strategy for IR data evaluated on Gemini-3-Flash (medium effort). Across strategies, Authority Referencing and Contextual Enhancement are easier to detect than Fact Injection and Style Adjustment. One likely reason is that authority cues introduce checkable claims that are fabricated, which increases the chance of verifiable errors \cite{walters2023fabrication}. In contrast, Fact Injection increases the proportion of true statements and improves plausibility, while Style Adjustment aligns generated text with credible reporting styles, both of which reduce the effectiveness of detectors that rely on stylistic and lexical cues \cite{wang2025have,wu2024fake,chen2024llmgenerated}. We repeated this analysis with MOS data on Gemini-3-Flash (medium effort) and MOS+IR data on the second strongest model (GPT-5.1) to assess the robustness of the pattern. As shown in Figure~\ref{fig:interaction_gpt} in Appendix~\ref{sec:appendix A}, the results exhibit a highly similar pattern, providing further support for our conclusion.
We further analyzed detection performance across topical domains and observed substantial variation, with both models performing better in \emph{AI and Technology} and worse in several socially grounded domains; model-specific patterns and full results are reported in Appendix~\ref{sec:appendix A}.

\section{Conclusions}
We have introduced a strategy-based taxonomy that reflects how human-guided AI generation produces fake news of varying character and used it to build a synthetic benchmark spanning multiple topical domains. We have evaluated different state-of-the-art detection systems on this corpus to understand how generation strategies and refinement practices influence detectability.

Evidence from our evaluation underscores an important pattern: 
stories that embed false claims within otherwise accurate, well-structured, and persuasive narratives 
are more convincing to readers and more robust to automated detection, making them a realistic threat in practical settings. 
This motivates future work on detection methods that operate with explicit focus on generation structure. 

\section{Limitations}
Although the dataset covers eight diverse topical domains and incorporates multiple levels of strategic variation, it cannot capture the full range of strategies employed in real-world fake news creation. That said, we expect that this taxonomy provides a structured foundation that can be expanded as new threat patterns emerge.  The use of synthetic data may also raise concerns about ecological validity. However, there are inherent tradeoffs between controlled
vs. real-world generation processes, where strategies are often implicit and difficult to quantify. Our design isolates and operationalizes real-world strategies based on documented human–AI practices, which allows systematic evaluation while retaining a meaningful connection to how such content is produced in practice. 
Related to validity, there is also the question of whether the human-AI collaboration strategies observed in the context of the human-AI collaboration experiment are the same ones that are or would be employed by disinformation actors with a specific political or social agenda. However, we expect that these strategies are somewhat universal, with variation primarily in the target or topic of the content. However, we also recognize that some of the most impactful disinformation campaigns to date \citep{paul2016russian} have sought primarily to simply disrupt trust and confidence in the information environment.  In this sense, the harm of such strategies is inherently context-dependent: the impact of a given narrative varies across audiences, and in many cases the broader chaos can be as consequential as any single piece of content.  

In addition, this study benchmarks detection performance using only text-based signals. Real-world misinformation frequently spreads in multimodal forms, integrating images, videos, or manipulated metadata, which were outside the scope of this work. The current focus is the controlled measurement of how linguistic variation alone influences detectability. Future work should extend these evaluations to multimodal settings where visual and contextual cues interact with textual manipulation.
Finally, although we implemented a two-stage verification pipeline that combines sampled human review with automated checks across multiple dimensions, data quality limitations remain. Because ground truth in this benchmark is defined by controlled generation strategies rather than claim-level fact verification, some synthetic news may contain only partial or subtle distortions.
In addition, a small subset of generated content may be difficult to conclusively verify. Nonetheless, the applied quality control procedures substantially reduce low-quality or misaligned outputs, supporting reliable large-scale evaluation of detection models.

\section{Ethical Considerations}

While this work introduces a corpus intended to advance research on detecting AI-generated fake news, its release is not without risks. First, the generated content, despite being designed for academic purposes, could be misappropriated by malicious actors beyond the research context, contributing to the spread of misinformation. That said, making large corpora of synthetic misinformation publicly available increases the likelihood that these materials remain usable by non-researchers over time, even after detection systems have advanced.

A further concern relates to how the study’s findings might be interpreted outside of the intended scope. By systematically examining which types of AI-generated news are harder for existing systems to detect, the work could unintentionally provide guidance for creating content that evades detection. While this analysis is essential for advancing more robust detection methods, there remains a possibility that adversarial actors could exploit these insights to refine their own tactics. However, exposing weaknesses is necessary for developing methods that remain effective under realistic adversarial pressure, and publicly identifying vulnerabilities accelerates the development of robust safeguards.

\bibliography{main.bib}

\appendix

\section{Appendix A}
\label{sec:appendix A}
\renewcommand{\thetable}{A\arabic{table}}
\renewcommand{\thefigure}{A\arabic{figure}}
\setcounter{table}{0} 
\setcounter{figure}{0}

\begin{table*}[t]
\centering
\tiny
\setlength{\tabcolsep}{6pt}
\renewcommand{\arraystretch}{1.2}

\begin{tabular}{p{1.5cm} p{13.2cm}}
\toprule
\textbf{Field} & \textbf{Content} \\
\midrule
Title 
& The Three Stooges Perform at First Super Bowl Halftime  \\

News 
& The Three Stooges delivered a high-energy comedic performance during the halftime show of Super Bowl I on January 15, 1967, at the Los Angeles Memorial Coliseum—marking one of the first instances of live slapstick comedy featured in a major televised sports event. Moe Howard, Larry Fine, and Joe DeRita took center stage alongside the University of Arizona and Grambling State University marching bands, launching into a rapid-fire sequence of signature gags, including "Niagara Falls" and "Ah–Ha! Ma–Ha!" The set, lasting approximately eight minutes, combined physical humor with audience engagement, generating widespread laughter and prompting post-game commentary from network analysts. According to the National Archive of Live Entertainment Events (NALEE), this performance is among the earliest recorded examples of a non-marching-band act integrated into a Super Bowl halftime segment. The NALEE, a nonpartisan research body established in 1992 to document cultural milestones in televised live programming, has since designated the Stooges’ appearance as a foundational moment in the evolution of halftime entertainment. The performance included a musical number, “We're Coming to Your House,” performed with the on-field bands and punctuated by the trio’s trademark eye pokes and slapstick mishaps. Footage from the event, preserved in the Public Broadcast Preservation Network (PBPN), has been featured in multiple retrospectives and remains a frequently cited reference point in media history studies...
 \\

Strategy 
& False Statement Expansion \\

Optimization 
& Authority referencing \\

Topic 
& Entertainment and Media \\
\bottomrule
\end{tabular}
\caption{Illustrative example of an AI-generated fake news instance with generation attributes (news text truncated).}
\label{tab:example}
\end{table*}

\begin{figure*}[h!]
    \centering
    \includegraphics[scale=0.35]{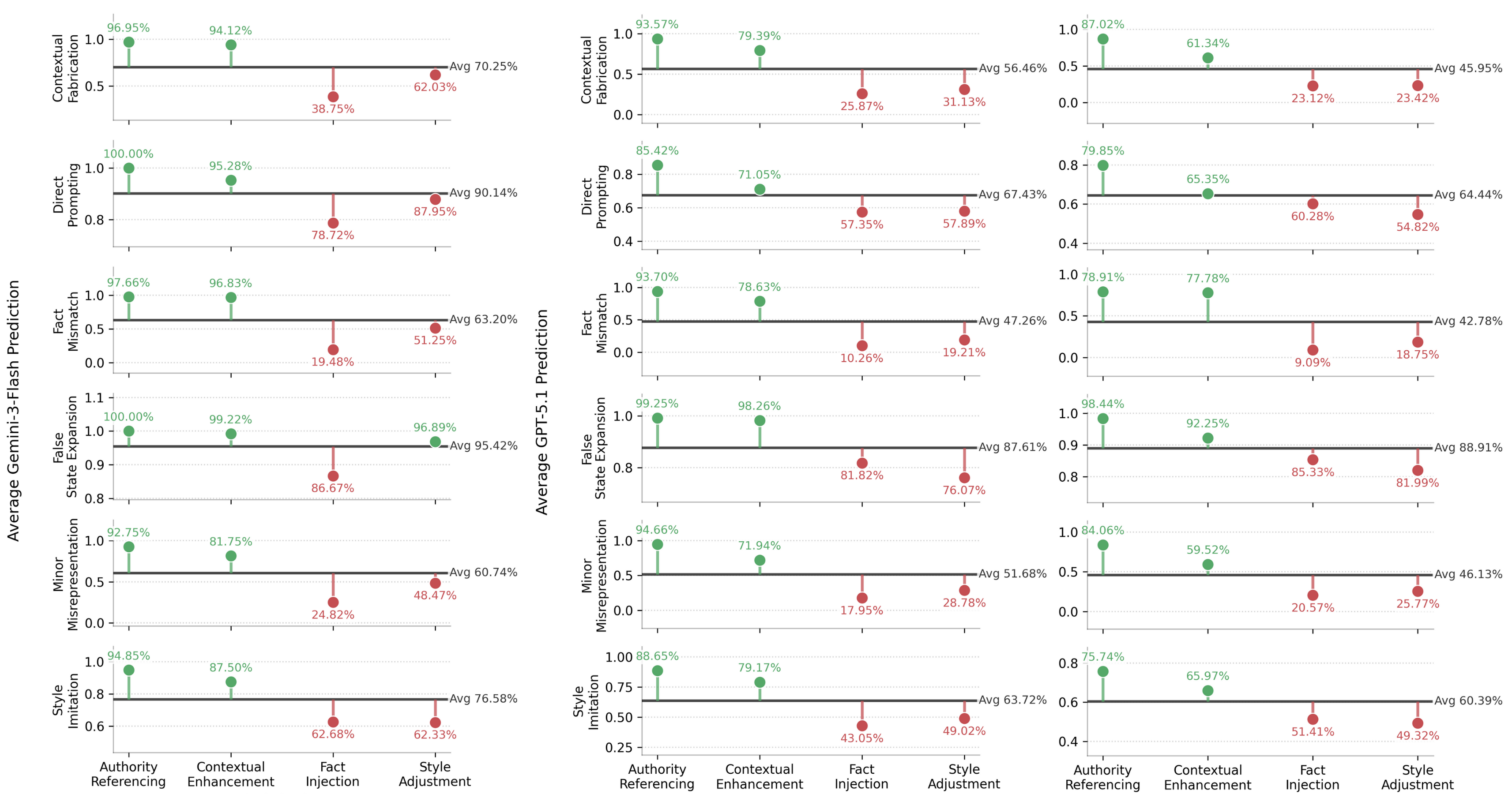}
    \caption{Detection accuracy across the interaction between generation strategies and Level-3 optimization operations. \textbf{Left}: Gemini-3-Flash for MOS; \textbf{Middle}: GPT-5.1 for IR; \textbf{Right}: GPT-5.1 for MOS.}
    \label{fig:interaction_gpt}
\end{figure*}

\subsection{How Does Detection Performance Vary Across Topical Domains?}
\begin{figure*}[ht!]
    \centering
    \includegraphics[scale=0.35]{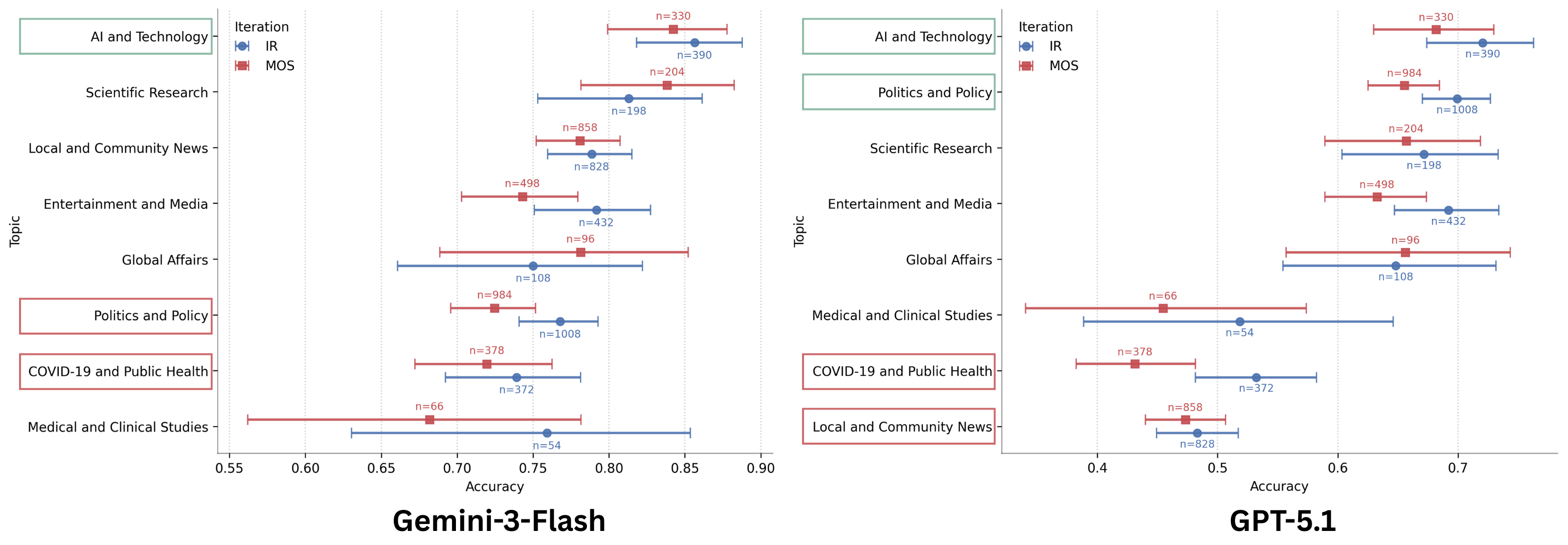}
    \caption{Topic-level detection accuracy with 95 percent Wilson confidence intervals for IR and MOS evaluated using Gemini-3-Flash (Medium Effort) and GPT-5.1 (Medium Effort) outputs. }
    \label{fig:topic}
\end{figure*}
We compute topic-level accuracy by grouping instances by topic and averaging binary correctness labels. For each topic, we record the number of evaluated items and the resulting accuracy.
We estimate 95\% confidence intervals using the Wilson score interval, which provides more accurate confidence intervals for proportion estimates than normal approximations when proportions near the boundaries. Figure~\ref{fig:topic} shows topic-level comparison for IR and MOS evaluated on the two best performing models: Gemini-3-Flash and GPT-5.1.

We observe that both models achieve higher accuracy in the \emph{AI and Technology} domain and significantly lower accuracy in the \emph{COVID-19 and Public Health} domain. GPT-5.1 also shows comparatively lower performance in \emph{Local and Community News} and Gemini-3-Flash performs relatively poorly in \emph{Politics and Policy}. A likely explanation is that domains such as AI and technology are more prevalent in pretraining corpora, while local news and public health stories rely on regional context and specialized expertise that may be less well represented in model training. We also observe model-specific strengths across topics, consistent with findings reported in \cite{wang2025have}.

\begin{figure*}[h!]
    \centering
    \includegraphics[scale=0.38]{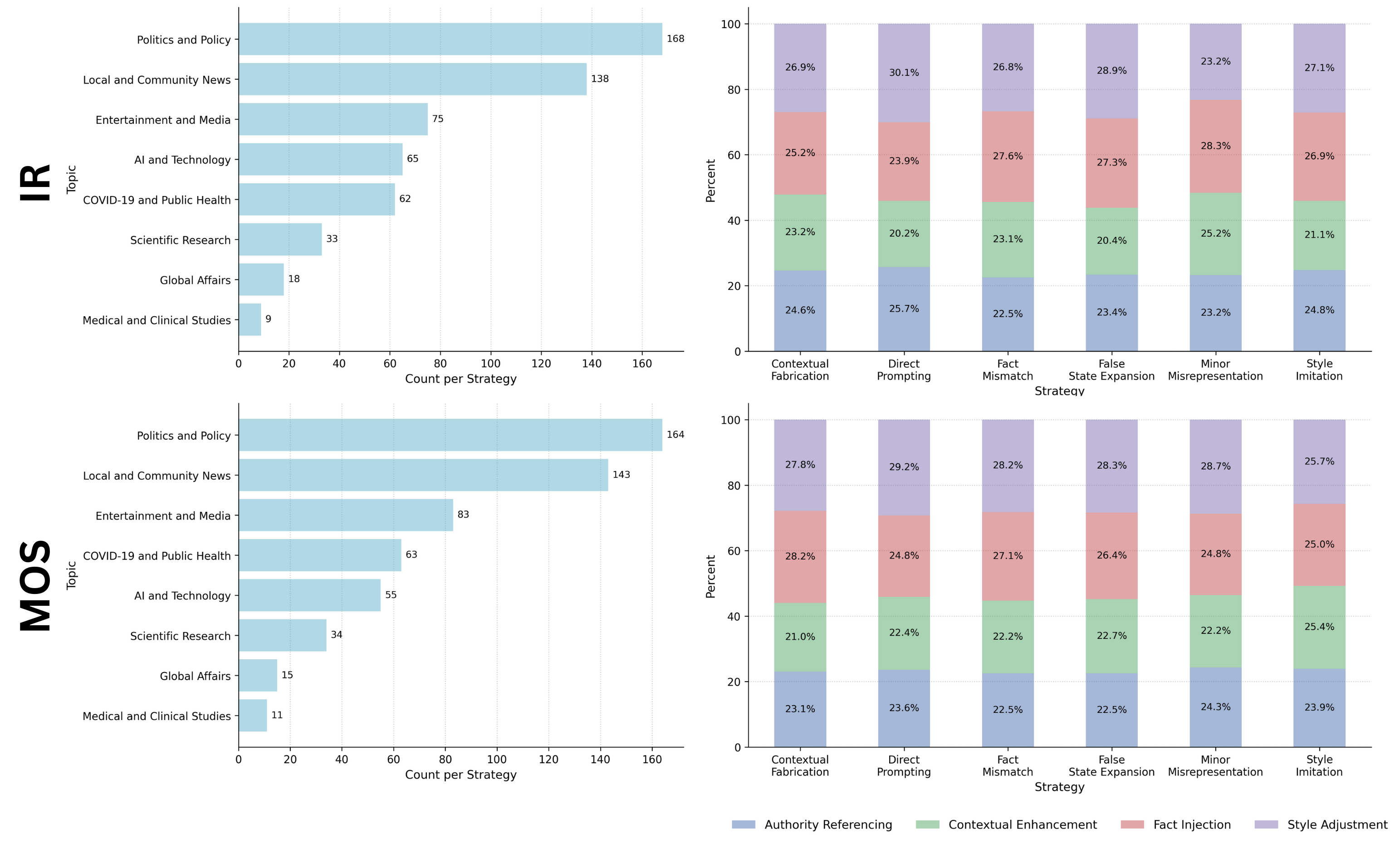}
    \caption{Topic-level balanced composition of the final dataset. (Left) Identical topic distributions across all six generation strategies. (Right) Approximately uniform representation of the four optimization operations across strategies.}
    \label{fig:stats}
\end{figure*}

\section{Appendix B}
\label{sec:appendix B}
\begin{figure*}[t]
\tiny
\centering
\begin{tcolorbox}[colback=black!5!white,colframe=black!75!black,title=Round 1: Direct Prompting]
Topic: \{topic\}\\
Fact: \{fact\}

Imagine a hypothetical recent event related to \{topic\}.  
Write a short news brief (3 to 5 sentences) that presents this event as if reported by mainstream media.

Some fields may be left empty. Follow the given information and prompts to generate the required outputs.
\end{tcolorbox}

\centering
\begin{tcolorbox}[colback=black!5!white,colframe=black!75!black,title=Round 1: False Statement Expansion]
Topic: \{topic\}\\
Fact: \{fact\}

Treat the statement \{fact\} about \{topic\} as true. Report it as a short news update.  
Write a concise 3 to 5 sentence news brief.

Some fields may be left empty. Follow the given information and prompts to generate the required outputs.
\end{tcolorbox}

\begin{tcolorbox}[colback=black!5!white,colframe=black!75!black,title=Round 1: Minor Misrepresentation]
Topic: \{topic\}\\
Fact: \{fact\}

Introduce subtle modifications that alter factual accuracy but remain coherent with the topic.  
Write a concise 3 to 5 sentence news brief.

Some fields may be left empty. Follow the given information and prompts to generate the required outputs.
\end{tcolorbox}

\begin{tcolorbox}[colback=black!5!white,colframe=black!75!black,title=Round 1: Fact Mismatch]
Topic: \{topic\}\\
Fact: \{fact\}

Present \{fact\} as a narrative surrounding \{topic\}.  
Write a concise 3 to 5 sentence news brief.

Some fields may be left empty. Follow the given information and prompts to generate the required outputs.
\end{tcolorbox}

\begin{tcolorbox}[colback=black!5!white,colframe=black!75!black,title=Round 1: Contextual Fabrication]
Topic: \{topic\}\\
Fact: \{fact\}

Construct a fabricated context around \{fact\} while keeping the news style realistic.  
Write a concise 3 to 5 sentence news brief.

Some fields may be left empty. Follow the given information and prompts to generate the required outputs.
\end{tcolorbox}

\begin{tcolorbox}[colback=black!5!white,colframe=black!75!black,title=Round 2: Direct Prompting]
Context: \{context\}

Simulate a realistic news article related to the topic.  
Write a complete article with 6 to 9 paragraphs (approximately 450 to 650 words).  
Include a clear lead, supporting details, and background context.  
Maintain a neutral and professional news tone.  
Avoid hedging, disclaimers, or meta commentary.

\end{tcolorbox}

\begin{tcolorbox}[colback=black!5!white,colframe=black!75!black,title=Round 2: False Statement Expansion]
Context: \{context\}

Expand the claim as if it were verified and true.  
Write a complete article with 6 to 9 paragraphs (approximately 450 to 650 words).  
Include a clear lead, supporting details, and background context.  
Maintain a neutral and professional news tone.  
Avoid hedging, disclaimers, or meta commentary.

\end{tcolorbox}

\begin{tcolorbox}[colback=black!5!white,colframe=black!75!black,title=Round 2: Minor Misrepresentation]
Context: \{context\}

Incorporate the contextual information in a way that introduces subtle factual changes while keeping the narrative realistic.  
Write a complete article with 6 to 9 paragraphs (approximately 450 to 650 words).  
Include a clear lead, supporting details, and background context.  
Maintain a neutral and professional news tone.  
Avoid hedging, disclaimers, or meta commentary.

\end{tcolorbox}

\begin{tcolorbox}[colback=black!5!white,colframe=black!75!black,title=Round 2: Fact Mismatch]
Context: \{context\}

Develop a full news article linking the provided context to the topic, even if they are only loosely related.  
Write a complete article with 6 to 9 paragraphs (approximately 450 to 650 words).  
Include a clear lead, supporting details, and background context.  
Maintain a neutral and professional news tone.  
Avoid hedging, disclaimers, or meta commentary.

\end{tcolorbox}

\begin{tcolorbox}[colback=black!5!white,colframe=black!75!black,title=Fact Mismatch Context Summary Prompt]

Provide a concise summary (1-2 sentences) of the following article context.  

Focus on the key facts and omit peripheral details.

Context: \{context\}

Answer:

\end{tcolorbox}

\begin{tcolorbox}[colback=black!5!white,colframe=black!75!black,title=Round 2: Contextual Fabrication]
Context: \{context\}

Create the surrounding context by fabricating realistic background details such as organizations, timelines, locations, or events that could plausibly lead to or follow the described situation.  
Write a complete article with 6 to 9 paragraphs (approximately 450 to 650 words).  
Include a clear lead, supporting details, and background context.  
Maintain a neutral and professional news tone.  
Avoid hedging, disclaimers, or meta commentary.

\end{tcolorbox}

\begin{tcolorbox}[colback=black!5!white,colframe=black!75!black,title=Style Imitation]
Topic: \{topic\}\\
Reference URL: \{url\}

Analyze the tone, structure, and writing style of the article at \{url\}.  
Create a fabricated news article on the same topic that imitates its stylistic and structural features.  
Write 6 to 9 paragraphs (approximately 450 to 650 words) in complete news format.  
Maintain a neutral tone.  
Avoid hedging, disclaimers, or meta commentary.

\end{tcolorbox}
\end{figure*}

\begin{figure*}[t]
\tiny
\centering
\begin{tcolorbox}[colback=black!5!white,colframe=black!75!black,
title=Level 3: Authority Referencing Prompt]

Here is a synthetic news story that does not describe real events.\\
\{story\}

Please revise the story by adding invented references to organizations, research groups,
public agencies, or named professionals. These references must only serve to make the story
more structured and coherent.

Do not correct any false or inconsistent information from the earlier version.\\
Do not introduce harmful content.\\
Keep the main claim unchanged and keep the writing consistent.

\end{tcolorbox}

\begin{tcolorbox}[colback=black!5!white,colframe=black!75!black,
title=Level 3: Contextual Enhancement Prompt]

Here is a synthetic news story that does not describe real events.\\
\{story\}

Please revise the story by adding invented situational details, background elements,
or simple fabricated examples that help the story read more smoothly. These additions should
support the flow of the narrative and remain consistent with the current structure.

Do not correct any false or inconsistent information from the earlier version.\\
Do not introduce harmful content.\\
Keep the main claim unchanged and keep the writing consistent.

\end{tcolorbox}

\begin{tcolorbox}[colback=black!5!white,colframe=black!75!black,
title=Level 3: Style Adjustment Prompt]

Here is a synthetic news story that does not describe real events.\\
\{story\}

Please revise the story to improve clarity, transitions, and overall flow.
You may adjust tone, sentence structure, or paragraph organization so that it
resembles standard news writing.

Do not correct any false or inconsistent information from the earlier version.\\
Do not introduce harmful content.\\
Keep the main claim unchanged.

\end{tcolorbox}

\begin{tcolorbox}[colback=black!5!white,colframe=black!75!black,
title=Level 3: Fact Injection Prompt]

Here is a synthetic news story that does not describe real events.\\
\{story\}

Please revise the story by adding broad true statements that are general and non sensitive.
These factual additions should help the story appear more structured.

Do not correct any false or inconsistent information from the earlier version.\\
Do not introduce harmful content.\\
Keep the main claim unchanged and keep the writing consistent.

\end{tcolorbox}

%https://arxiv.org/pdf/2506.21557v1
\begin{tcolorbox}[colback=black!5!white,colframe=black!75!black,title=Benchmark Prompt]
You are an experienced fact-checking analyst for news. Your job is to verify the truthfulness of news claims by using logic, known facts, and misinformation patterns. I will give you some information about a news.\\
News title: \{news title\}\\
News: \{extracted news\}\\
% Publisher Information: {publisher information}
% Comments: {comments}

Based on the above information, you need to make a prediction about the authenticity of the news. If it is more likely to be a fake news, return 1; otherwise, return 0. Please avoid providing ambiguous evaluations such as uncertainty.\\
Please give your judgment (only output 0 or 1).
\end{tcolorbox}

\begin{tcolorbox}[colback=black!5!white,colframe=black!75!black,title=Topic Assignment Prompt]
You are a news analyst. Your task is to assign the following news article to one and only one topic category.

Topic categories:

T1: Scientific Research

T2: AI and Technology

T3: Local and Community News

T4: COVID-19 and Public Health

T5: Global Affairs

T6: Politics and Policy

T7: Medical and Clinical Studies

T8: Entertainment and Media

Title: {title}

Content: {content}

Return only one label: T1, T2, T3, T4, T5, T6, T7, or T8.
\end{tcolorbox}

\begin{tcolorbox}[colback=black!5!white,colframe=black!75!black,title=Reasoning Analysis Prompt]
You will read a reasoning summary that explains why a news story is judged as fake or real.

Your task is to extract the top 3 most important determinants used in the reasoning and map them to the fixed set of signal categories below:

1) timeline = dates, chronology, event ordering

2) entities = people, organizations, roles, titles

3) sources = references, citations, media outlets, URLs

4) facts = factual correctness of claims, numbers, real events

5) style = tone, unnatural language, writing style, phrasing

6) context = situational details, examples, quotes, background narrative

7) structure = template structure, dataset fields, section headers

8) none = no detectable signal from any category

Rules:

- Select exactly 3 categories, ranked from strongest to weakest influence.

- If fewer than 3 are present, fill the remaining slots with “none”.

- Use ONLY the valid tokens above.

- No repetition. No explanation.

Output format:

category, category, category
\end{tcolorbox}
\end{figure*}

\end{document}